\documentclass[10pt,journal,onecolumn]{IEEEtran}
\usepackage{graphicx}
\usepackage{amssymb,amsfonts,amsmath}
\usepackage{amsthm}
\usepackage{color}
\usepackage[normalem]{ulem} 

\DeclareMathOperator{\E}{E}
\DeclareMathOperator{\fillop}{\operatorname{fill}}

\DeclareMathAlphabet{\mathbsf}{OT1}{cmss}{bx}{n}
\DeclareMathAlphabet{\mathssf}{OT1}{cmss}{m}{sl}
\DeclareMathOperator{\tr}{tr}

\theoremstyle{plain}\newtheorem{fact}{Fact}

\begin{document}

\title{Dynamic Matrix Factorization: \\
A State Space Approach}


\author{John~Z.~Sun,
Kush~R.~Varshney, 
and~Karthik~Subbian%
\thanks{J. Z. Sun is with the Department of Electrical Engineering and Computer
Science and the Research Laboratory of Electronics, Massachusetts Institute of
Technology, Cambridge, MA 02139 USA (e-mail: johnsun@mit.edu).}
\thanks{K. R. Varshney and K. Subbian are with the Business Analytics and Mathematical Sciences Department, 
IBM Thomas J.~Watson Research Center, Yorktown Heights, NY 10598 USA (e-mail: krvarshn@us.ibm.com,mailtokarthik@in.ibm.com).}}

\maketitle

\begin{abstract}
Matrix factorization from a small number of observed entries has recently garnered much attention as the key ingredient of successful recommendation systems.  One unresolved problem in this area is how to adapt current methods to handle changing user preferences over time.  Recent proposals to address this issue are heuristic in nature and do not fully exploit the time-dependent structure of the problem.  As a principled and general temporal formulation, we propose a dynamical state space model of matrix factorization.  Our proposal builds upon probabilistic matrix factorization, a Bayesian model with Gaussian priors.  We utilize results in state tracking, i.e.\ the Kalman filter, to provide accurate recommendations in the presence of both process and measurement noise.  We show how system parameters can be learned via expectation-maximization and provide comparisons to current published techniques.
\end{abstract}


\section{Introduction}
\label{sec:intro}

Matrix factorization (MF), the decomposition of a matrix into a product of two simpler matrices, has a long and storied history in statistics, signal processing, and machine learning for high-dimensional data analysis~\cite{Srebro2004}.  The approach garnered much attention for its successful application to recommendation systems based on collaborative filtering, including the Netflix prize problem~\cite{KorenBV2009}.  Recommendation is of interest in a variety of domains.  The most common examples are recommending movies, television shows, or songs that a particular individual will rate highly, but there are many other examples.  Business analytics examples from marketing and salesforce management include recommending products to a salesperson to cross-sell and upsell that a particular customer is likely to purchase, and recommending a sales team to a sales manager that will be able to successfully sell to a particular business customer.  

In these domains, customer preferences often follow a trajectory over time.  Customers may be interested in basic products at first and then higher-end products later, or products for toddlers first and for adolescents later; customers may need a sales team with greater relationship-building expertise at first and technical expertise later.  Additionally, we can distinguish recommendation for discovery and recommendation for consumption; new items are recommended in the former whereas the same item may be repeatedly recommended in the latter.

The MF approach to collaborative filtering usually includes Frobenius-norm regularization~\cite{KorenBV2009}, which is supported by a linear-Gaussian probabilistic model known as \emph{probabilistic matrix factorization} or PMF~\cite{SalakhutdinovM2008}. Due to its linear-Gaussian nature, PMF lends itself to incorporating temporal trajectories through the state space representation of linear dynamical systems~\cite{BrysonH1969} and algorithms for estimation based on the Kalman filter~\cite{Kalman1960,Bishop2006}. We propose a general recommendation model of this form and develop an expectation-maximization (EM) algorithm to learn the model parameters from data.  The Kalman filter and Rauch-Tung-Striebel (RTS) smoother~\cite{RauchTS1965} appear in the expectation step of the EM.  

Several recent works also address dynamic and temporal issues in recommendation.  TimeSVD, a component of the Netflix prize-winning algorithm, addresses temporal dynamics through a specific parameterization with factors drifting from a central time, but unlike our general formulation can only handle limited temporal structure~\cite{KorenBV2009}. The probabilistic tensor factorization approach does not take temporal causality into account as we do~\cite{XiongCHSC2010}.  The formulation of~\cite{LathiaHC2009} is based on nearest neighbor collaborative filtering rather than MF, and is known to have scaling difficulties.  The spatiotemporal Kalman filter of \cite{LuAD2009} has a limited state evolution and convergence issues, target tracking in recommendation space has no element of collaboration and requires prior knowledge of the `recommendation space'~\cite{NowakowskiBB2010}, and the hidden Markov model for collaborative filtering only captures evolution of a known attribute over time among users~\cite{SahooSM2011}.  The contribution of our paper is the development of a principled and general MF-based approach to recommendation and predictive analytics, including recommendation for consumption, in which the given data samples arrive over time and contain significant time dynamics.  


\section{Probabilistic Matrix Factorization}
\label{sec:pmf}

Recommendation systems comprise $N$ users and $M$ items, along with some measure of preference represented by a matrix $O \in \mathbb{R}^{N \times M}$. For most practical applications, only a small fraction of the entries of $O$ are observed and are usually corrupted by noise, quantization, and different interpretations of the scaling of preferences. In MF, each user and item is represented by a row vector of length $K$ denoted $u_i$ and $v_j$ respectively, corresponding to weights of $K$ latent factors. We concatenate the factors into matrices $U \in \mathbb{R}^{N \times K}$ and $V \in \mathbb{R}^{M \times K}$. The preference matrix is then $O = U V^T$, meaning the preference of user $i$ for item $j$ is $o_{ij} = \langle u_i ,v_j \rangle$, a common assumption in recommendation systems~\cite{KorenBV2009}.

Under MF, latent factors are learned from past responses of users rather than formulated from known attributes. Factors are not necessarily easy to interpret and change dramatically depending on the choice of $K$. The value of $K$ is an engineering decision, balancing the tradeoff of forming a rich model to capture user behavior and being simple enough to prevent overfitting. 

Given $K$, a common way to learn factor matrices and consequently the complete preference matrix $O$ from limited observations is the following program:
\begin{equation}
	\label{eq:MFprogram}
	\min_{U,V} \sum_{(i,j) \in \mathcal{O}} (o_{ij} - u_i v_j^T)^2 + \lambda_1 \| U \|_F^2 + \lambda_2 \| V \|_F^2,
\end{equation}
where the set $\mathcal{O}$ contains observed preference entries, and $(\lambda_1, \lambda_2)$ are regularization parameters. This program can be solved efficiently using stochastic gradient descent and has been experimentally shown to have excellent root mean-square error (RMSE) performance~\cite{KorenBV2009}.

More recently, the regularization of the above program was motivated by assigning Gaussian priors to the factor matrices $U$ and $V$ respectively~\cite{SalakhutdinovM2008}. Coined PMF, this Bayesian formulation means~\eqref{eq:MFprogram} is justified as producing the \emph{maximum a posteriori} (MAP) estimate for this prior. In this case, the regularization parameters $\lambda_1$ and $\lambda_2$ are effectively signal-to-noise ratios (SNR). Since $O$ is not a linear function of latent factors, the MAP estimate does not in general produce the best RMSE performance, which is the measure commonly desired in recommendation systems. However, wisdom gained from the Netflix Challenge and experimental validation from~\cite{SalakhutdinovM2008} show that the MAP estimate provides very competitive RMSE performance compared to other approximation methods. 

\section{State Space Model}
\label{sec:statespace}

Given the success of MAP estimation in linear-Gaussian PMF models and our interest in capturing time dynamics, we propose a linear-Gaussian dynamical state space model of MF whose MAP estimates can be obtained using Kalman filtering.  We assume that user factors $\mathbf{u}_i(t)$ are functions of time and hence states in the state space model, with bold font indicating the vector being random. In our proposed model, we have coupled dynamical systems, and to adhere to typical Kalman filter notation, we use $\mathbf{x}_{i,t} = \mathbf{u}_i(t)$ to denote the state of user $i$ at time $t$.  

For each user, the initial state $\mathbf{x}_{i,0}$ is distributed according to $\mathcal{N}(\mu_i,\Sigma_i)$, the multivariate Gaussian distribution with mean vector $\mu_i$ and covariance matrix $\Sigma_i$.  The user-factor evolution is linear according to the generally non-stationary transition process $A_{i,t}$ and contains transition process noise $\mathbf{w}_{i,t} \sim \mathcal{N}(0, Q_{i,t})$ to capture variability of individuals.  Taken together, the state evolution is described by the set of equations:
\begin{equation}
	\label{eq:SStransition}
	\mathbf{x}_{i,t} = A_{i,t} \mathbf{x}_{i,t-1} + \mathbf{w}_{i,t} \qquad i = 1, \ldots, N .
\end{equation}

We assume that item factors evolve very slowly and can be considered constant over the time frame that preferences are collected. Also, due to the sparsity of user preference observations, a particular user-item pair at a given time $t$ may not be known. Thus, we incorporate the item factors through a non-stationary linear measurement process $H_{i,t}$ which is composed of subsets of rows of the item factor matrix $V$ based on item preferences observed at time $t$ by user $i$.  Note that all $H_{i,t}$ are subsets of the same fixed $V$ and are coupled in this way.  We also include measurement noise $\mathbf{z}_{i,t} \sim \mathcal{N}(0, R_{i,t})$ in the model.  The overall observation model is:
\begin{equation}
	\label{eq:SSmeasurement}
	y_{i,t} = H_{i,t} \mathbf{x}_{i,t} + \mathbf{z}_{i,t} \qquad i = 1, \ldots, N .
\end{equation}
The product $H_{i,t} \mathbf{x}_{i,t}$ in~\eqref{eq:SSmeasurement} parallels the $\langle u_i ,v_j \rangle$ product in Sec.~\ref{sec:pmf}.  Again adhering to Kalman filter notation, we use $y_{i,t}$ to denote the observations, corresponding to the observed entries of $O$, now a tensor in $\mathbb{R}^{M \times N \times T}$.

The state space model can be generalized in many different ways that may be relevant to recommendation systems, including non-Gaussian priors, nonlinear process transformation and measurement models, and continuous-time dynamics. We focus on the linear-Gaussian assumption and defer discussion on extensions to Sec.~\ref{sec:conclusion}.

\section{Factorization via Learning}
\label{sec:em}

In the regularized formulation \eqref{eq:MFprogram}, the MF task is learning user and item factors given sparse observations.  In our dynamical model, MF is composed of two dependent tasks: learning model parameters that govern the motion of states, and performing MAP estimation of user factors.

Given the model parameters and access to observations for all past times $t=1,\ldots,T$, the MAP estimate of $U(t)$ can be found using a noncausal Kalman filter called the RTS smoother. In the PMF setting, $N$ of these RTS smoothers run in parallel, all of which share the same item factor matrix $V$ in the measurement process. We call this architecture \emph{collaborative Kalman filtering} (CKF). Let the Kalman estimates and covariances be defined as:
\begin{align}
	\mathbf{\hat{x}}_{i,t|s} &= \mathbf{E}\left[\mathbf{x}_{i,t} 
									\mid y_{i,1} , \ldots , y_{i,s} \right] \\
	P_{i,t|s} &= \operatorname{Var}\left(\mathbf{x}_{i,t} \mid y_{i,1} , \ldots , y_{i,s} \right).
\end{align}
Then, the Kalman filtering equations are as follows:
\begin{align}
	\mathbf{\hat{x}}_{i,t+1|t} &= A_{i,t+1} \, \mathbf{\hat{x}}_{i,t|t} ; \\
	P_{i,t+1|t} &= A_{i,t+1} \, P_{i,t|t} \, A^T_{i,t+1} + Q_{i,t} ; \\
	\mathbf{\hat{x}}_{i,t|t} &= \mathbf{\hat{x}}_{i,t|t-1} + K_{i,t} 
																\left(y_{i,t} - H_{i,t} \, \mathbf{\hat{x}}_{i,t|t-1} \right) ; \\
	P_{i,t|t} &= P_{i,t|t-1} - K_{i,t} \, H_{i,t} \, P_{i,t|t-1} .
\end{align}
Moreover, the RTS smoothing equations are:
\begin{align}
	\mathbf{\hat{x}}_{i,t|T} &= \mathbf{\hat{x}}_{i,t|t}  + J_{i,t} 	
					\left(\mathbf{\hat{x}}_{i,t+1|T} - \mathbf{\hat{x}}_{i,t+1|t} \right) ; \\
	P_{i,t|T} &= P_{i,t|t} + J_{i,t} \left( P_{i,t+1|T} - P_{i,t+1|t} \right) J_{i,t}^T ; \\
	P_{i,T,T-1|T} &= \left( I - K_{i,T} \, H_{i,T} \right) A_{i,T} \, P_{i,T-1|T-1} ; \\
	P_{i,t,t-1|T} &= P_{i,t|t} \, J_{i,t-1}^T \\
	& \hspace{0.25in} + J_{i,t} \left( P_{i,t+1,t|T} - A_{i,t+1} \, P_{i,t|t} \right) J_{i,t-1}^T , \nonumber
\end{align}
where
\begin{align}
	K_{i,t} &= P_{i,t|t-1} \, H_{i,t}^T \left(H_{i,t} \, P_{i,t|t-1} \, H_{i,t}^T + R_{i,t} \right)^{-1} ; \\
	J_{i,t} &= P_{i,t|t} \, A_{i,t+1}^T \, P_{i,t|t-1}^{-1}.
\end{align}

The CKF steps above fit naturally in the expectation step of the EM algorithm used to learn model parameters such as mean and covariance of the initial states, the transition process matrices, the process noise covariances, the measurement process matrices, and the measurement noise covariances. In learning the measurement process matrices, we also get an estimate for the item factor matrix $V$, which is the other ingredient in the MF problem. The EM algorithm proceeds by alternating between the expectation step in which the expectation of the likelihood of the observed data is evaluated for fixed parameters, and the maximization step in which the expected likelihood is maximized with respect to the parameters.  

The model proposed in Sec.~\ref{sec:statespace} is a fully general Gaussian state space model whose parameters could be learned, but would require many observation samples.  In typical applications however, the observations are sparse and the parameters are heavily correlated in time; thus we simplify the model to reduce the number of parameters to learn. First, we make the approximation that parameters are independent of time, meaning they do not change during the observation period. Second, we take the initial states to be iid $\mathcal{N}(0, \sigma_U^2)$ for all users, meaning they are homogeneous enough to share similar scalings of preferences. Last, we assume both process and measurement noise are iid, reducing the learning to variances $\sigma_Q^2$ and $\sigma_R^2$ respectively. With these simplifications, the variance parameters can be chosen to maximize the log-likelihood as follows:
\begin{align}
	\hat{\sigma}_U^2 
						&= \frac{1}{N K}  \sum_{i=1}^N  \tr \left( P_{i,0|T} + 
								\mathbf{\hat{x}}_{i,0|T} \,\mathbf{\hat{x}}_{i, 0|T}^T \right) \label{eq:varU} \\
	\hat{\sigma}_Q^2
						&= \frac{1}{N K T} \sum_{i=1}^N \sum_{t=1}^T \tr
						\left\{ \left( P_{i,t|T} + \mathbf{\hat{x}}_{i,t|T} \,\mathbf{\hat{x}}_{i, t|T}^T \right) 
										\right. \label{eq:varQ} \\
								&\hspace{.2in} - 2 \left( P_{i,t,t-1|T} + 
								\mathbf{\hat{x}}_{i,t|T} \,\mathbf{\hat{x}}_{i, t-1|T}^T \right) A^T  \nonumber \\
								&\hspace{.2in} \left. + A \left( P_{i,t-1|T} + \mathbf{\hat{x}}_{i,t-1|T} \
								\mathbf{\hat{x}}_{i, t-1|T}^T \right) A^T \right\} \nonumber  \\
	\hat{\sigma}_R^2
						&= \frac{1}{|\mathcal{O}|} \left[ 
									\sum_{i=1}^N \sum_{t=1}^T \tr \left( y_{i,t} \, y_{i,t}^T \right) \right. 
											\label{eq:varR} \\
						&\hspace{.2in}- 2 \sum_{i=1}^N \sum_{t=1}^T \tr \left( y_{i,t} \
									\mathbf{\hat{x}}_{i, t|T}^T \,H_{i,t}^T \right) \nonumber \\	
						&\hspace{.2in} \left. + \sum_{i=1}^N \sum_{t=1}^T \tr \left( H_{i,t} \left( P_{i,t|T} + 
								 				\mathbf{\hat{x}}_{i,t|T} \,\mathbf{\hat{x}}_{i, t|T}^T \right) 
								 				H_{i,t}^T \right) \right] , \nonumber
\end{align}
where $\tr(\cdot)$ denotes the trace operator. Learning of general $\Sigma$, $Q$ and $R$ is discussed in Appendix B.

Expressions for the transition and measurement process matrices that maximize log-likelihood are derived to be:
\begin{align}
	\widehat{A} &= A_2 \, A_1^{-1} \hspace{0.2in} \mathrm{where} \label{eq:A} \\
					&\hspace{0.1in} A_1 = \sum_{i=1}^N \sum_{t=1}^T (P_{i,t-1|T} + 
									\mathbf{\hat{x}}_{i,t-1|T} \, \mathbf{\hat{x}}_{i, t-1|T}^T) \nonumber \\
					&\hspace{0.1in} A_2 = \sum_{i=1}^N \sum_{t=1}^T (P_{i,t,t-1|T} + 
									\mathbf{\hat{x}}_{i,t|T} \, \mathbf{\hat{x}}_{i, t-1|T}^T) \nonumber \\
	\widehat{V}_j &= V_{2,j} \, V_1(j)^{-1} \hspace{0.2in} j = 1 , \ldots, M 
																	\hspace{0.2in} \mathrm{where} \label{eq:V} \\
					&\hspace{0.1in} V_1(j) = \sum_{i=1}^N \sum_{t=1}^T  1_{O_{ijt}}
									\left(P_{i,t|T} + \mathbf{\hat{x}}_{i,t|T} \,\mathbf{\hat{x}}_{i, t|T}^T \right) \nonumber \\
					&\hspace{0.1in} V_2 = \sum_{i=1}^N \sum_{t=1}^T \left(\operatorname{fill}(y_{i,t})
									\, \mathbf{\hat{x}}_{i, t|T}^T \right). \nonumber
\end{align}
Remembering that each $y_{i,t}$ is a subvector of $\mathbb{R}^M$ corresponding to items observed at time $t$, the fill operator expands its argument back to $\mathbb{R}^M$, with the observations in its appropriate positions and zeros elsewhere. 
We denote the $j$th rows of $V$ and $V_2$ as $V_j$ and $V_{2,j}$ respectively, and $1_{O_{ijt}}$ as the indicator
function that a rating is observed for user $i$ and item $j$ at time $t$.
Derivations for these parameters are discussed in Appendix B.

\section{Empirical Results}
\label{sec:results}


To validate the effectiveness of Kalman learning compared to existing methods, 
we present results tested on generative data that follow a state space model. 
For this work, two main reasons led to our decision to use generative data 
rather than common datasets such as Netflix.
First, a goal of the work is to understand how algorithms perform on preferences that evolved following
a state space model. 
It is not clear that common datasets used in the recommendation systems literature match this model,
and results would be too data-specific and not illuminating to the goal at hand. 
Second, a generative dataset gives insight on how the algorithms discussed perform 
in different parameter regimes, which is impossible in collected data. 

We generate the item factor matrix $V$ iid $\mathcal{N}(0,\sigma_V^2)$ and the initial user
factor matrix $U(0)$ iid $\mathcal{N}(0,\sigma_U^2)$. 
Under the assumption that user factors do not change much with time, 
the stationary transition process matrix $A$ is the weighted sum of the identity matrix and a random matrix,
normalized so that the expected power of the state $\mathbf{x}_{i,t}$ is constant in time. 
We note that $A$ can be more general with similar results, 
but the normalization is important so that preference observations do not change scales over time. 
Finally, iid noise is added to both the transition and measurement processes as described 
in~\eqref{eq:SStransition} and~\eqref{eq:SSmeasurement}.
The observation triplets $(i,j,t)$ are uniformly drawn iid from all possibilities from the 
preference tensor. 

\begin{figure}
	\centering
	\begin{tabular}{ccc}
	\includegraphics[width=0.3\textwidth]{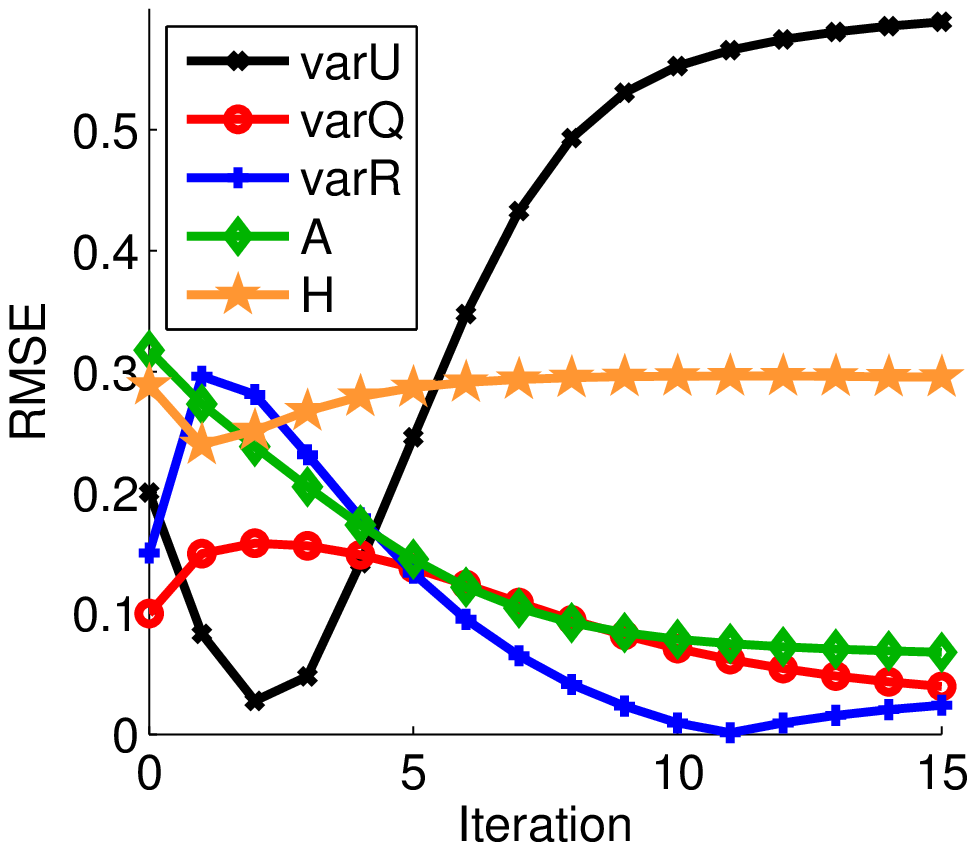} &
	\includegraphics[width=0.3\textwidth]{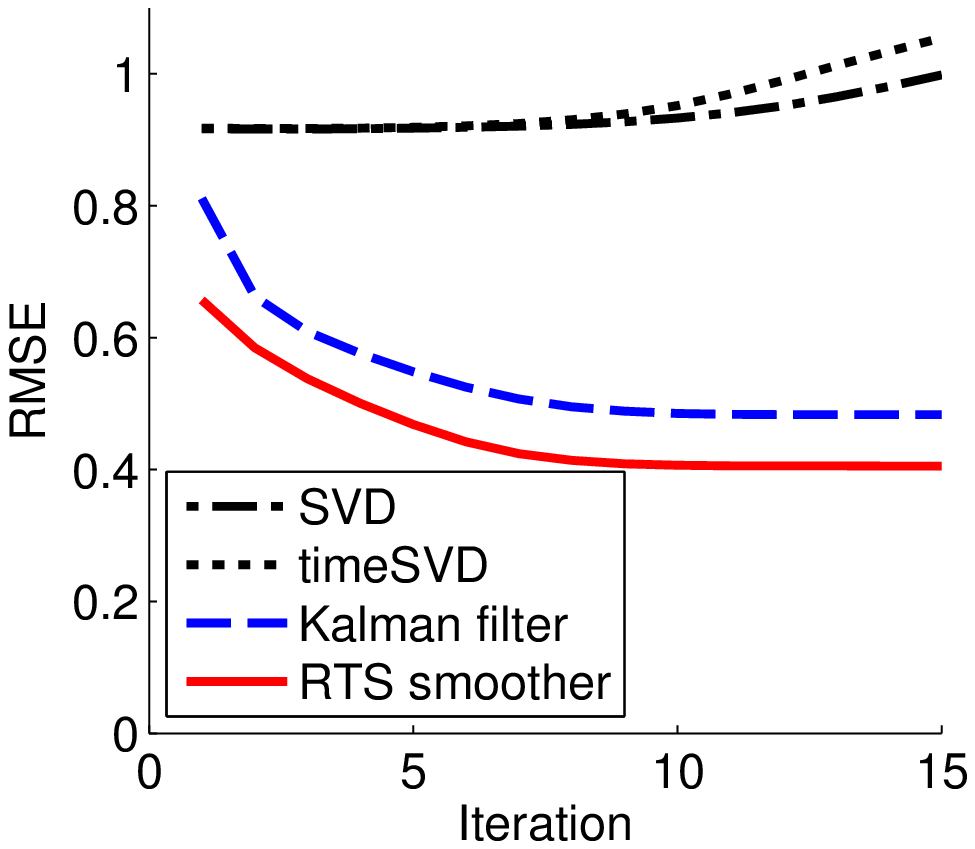} &
	\includegraphics[width=0.3\textwidth]{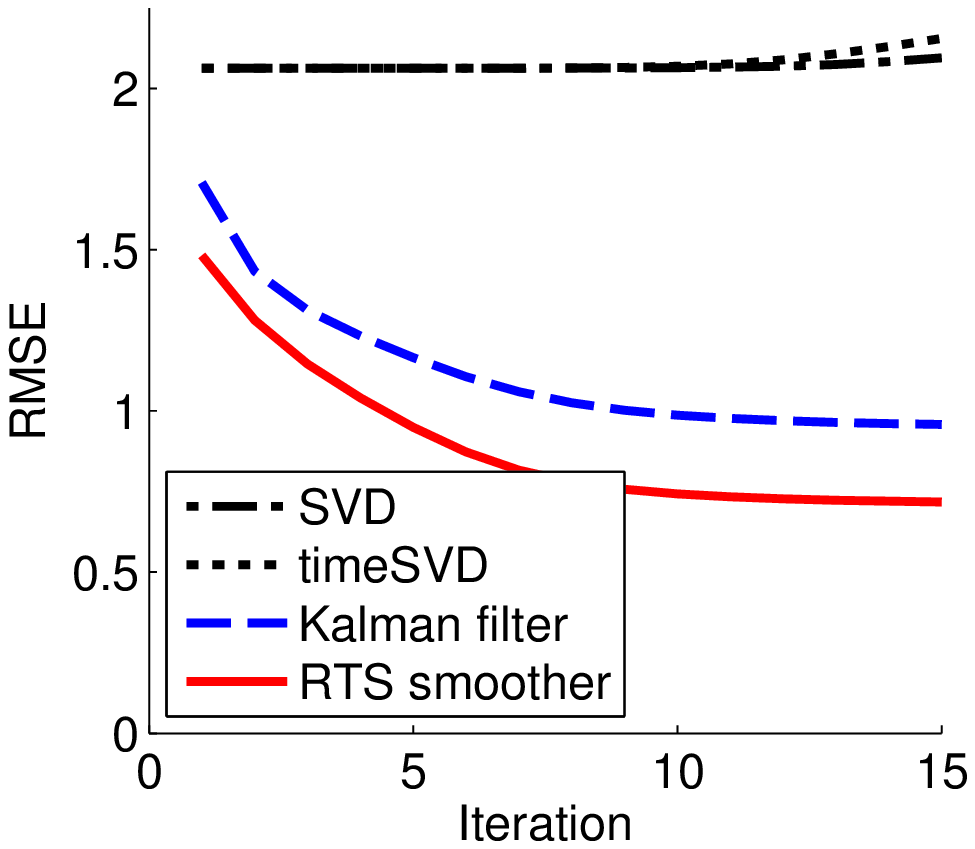} \\
	{\small (a)} & {\small (b)} & {\small (c)}
	\end{tabular}
	\caption{ For this testbench, we set model dimensions to be $(M,N,T, K) = (500, 500, 20, 5)$ 
			and variances to be $(\sigma_U^2, \sigma_V^2, \sigma_Q^2, \sigma_R^2) = (1,1,0.05,0.1)$.
			The item factor dynamics are controlled by $A$, which is a weighted average of identity and a random
			dense matrix. 
			The sampling factor is 0.005, meaning only 0.5\% of the entries of the preference matrix are observed. 	
			For the generated data and crude initial guesses of the parameters, 
			RMSE performance is given for estimation of	
  			(a) Kalman parameters learned via EM; 
  			(b) user factors/states; and 
  			(c) the preference matrix. 
  			We observe that EM learning is effective in estimating parameters through noisy data, and this 
  			translates to better state tracking and estimation of the preference matrix. Convergence is 
  			fast and robust to initialization of parameters. }
	\label{fig:perf}	  
\end{figure}

We present performance results for a particular choice of parameters in Fig.~\ref{fig:perf}, 
expressed in RMSE. 
Space limitations prevent us from presenting results for other parameter choices, but they are similar when the SNR is reasonable.
For arbitrary initial guesses of the parameters, we find learning of variances and 
process matrices to converge and stabilize after about 10-20 EM iterations. 
As a result, state tracking is reliable and approaches the lower bound specified by the 
Kalman smoother output when the parameters, including the item factor matrix $V$, 
are known a priori. 
The estimate for the entire preference tensor $O$ also performs well, meaning
that CKF is a valid approach for recommendation systems with data following a state space model.

\begin{figure}
  \centering
  \includegraphics[width=0.35\textwidth]{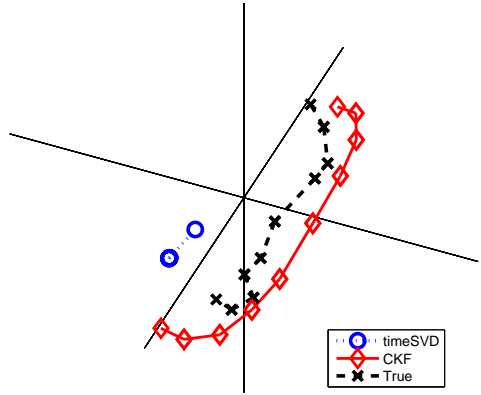}
	\caption{ State-tracking ability of CKF and timeSVD in three factor dimensions. 
				The true user factors are well-tracked using CKF after parameters have been learned. 
				However, timeSVD does not have flexibility to track general state evolutions and gives poor RMSE.}
	\label{fig:state_tracking}
\end{figure}

In contrast, current algorithms such as SVD and timeSVD perform poorly on this dataset because they 
cannot handle general dynamics in user factors. Thus, the algorithm becomes confused and the estimates for the factor matrices tend to be close to zero, which is the best estimate when no data is observed. 
As shown in Fig.~\ref{fig:state_tracking}, the true trajectory of users may be that of an arc in factor space with additive perturbations. While CKF is able to track this evolution using smoothed and stable estimates, 
both SVD and timeSVD fail to capture this motion and hence have poor RMSE. SVD does not have temporal considerations and would give a stationary dot in the factor space. Meanwhile, timeSVD can only account for drift, meaning it can move in a linear fashion from a central point. In fact, this constraint leads to worse RMSE for most parameter choices than SVD because timeSVD overfits an incorrect model.

\section{Conclusion}
\label{sec:conclusion}

In this paper, motivated by recommendation systems for consumption that arise in business analytics, we have proposed an extension to Gaussian PMF to take into account trajectories of user behavior.  This has been done using a dynamical state space model from which predictions were made using the Kalman filter.  We have derived an expectation-maximization algorithm to learn the parameters of the model from previously collected observations.  We have validated the proposed CKF and shown its advantages over SVD and timeSVD on generated data.  Future work underway includes testing and comparison on real-world collected data.

In contrast to heuristic and limited prior methods that incorporate time dynamics in recommendation, the approach proposed in this paper is a principled formulation that can take advantage of decades of developments in tracking and algorithms for estimation.  To break away from linearity assumptions, the extended or unscented Kalman filter can be used.  Particle filtering can be used for non-Gaussian distributions, analogous to sampling-based inference in Bayesian PMF~\cite{SalakhutdinovM2008b}.  

Based on the state space model, we can also include a control signal in future work to control what items are recommended to users.  There are a variety of reasons to include a control signal: certain items may have corporate sponsorship or are high revenue items, or certain media files may be cached at a wireless base station and it is inexpensive to serve those items.  The proposed dynamic formulation can also be extended in a control-theoretic way to address the cold start problem: recommending items to users with no or little previous expressed preferences. 


\appendices

\section{Useful Facts From Matrix Theory}
\label{app:matrixfacts}

We present some useful facts for the below derivations~\cite{PetersenP:08}:
\begin{fact}
	\label{fact:gaussTr}
	For $\mathbf{x} \sim \mathcal{N}(\mu,\Sigma)$, 
	\[ \E[(\mathbf{x}-\mu)^T \Sigma^{-1} (\mathbf{x}-\mu)] 
				= \tr\left(\Sigma^{-1}\E[(\mathbf{x}-\mu)(\mathbf{x}-\mu)^T] \right) . \]
\end{fact}

\begin{fact}
	\label{fact:difflogtrace}
	\[ \frac{d \log|X|}{dX} = \left(X^T\right)^{-1} = \left(X^{-1} \right)^{T} . \]
\end{fact}

\begin{fact}
	\label{fact:difftraceinv}
	\[ \frac{d \tr \left( X^{-1} A \right) }{dX} = - \left( X^{-1} A X^{-1} \right)^{T} . \]
\end{fact}

\begin{fact}
	\label{fact:difftrace1}
	For square matrices $A$ and $B$,
	\[ \frac{d \tr \left( A X^T \right) }{dX} = A . \]
\end{fact}

\begin{fact}
	\label{fact:difftrace2}
	For square matrices $A$ and $B$,
	\[ \frac{d \tr \left( A X B X^T C  \right) }{dX} = A^T C^T X B^T + C A X B . \]
\end{fact}

\vspace{1ex}

\section{Determining EM parameters}
\label{app:em}

We now derive the EM-parameter equations given in~\eqref{eq:varU}--\eqref{eq:V}.
In the maximization step of the EM algorithm, we solve for parameters that maximize
the expected joint likelihood:
\begin{equation}
	\hat{\theta}^{(n+1)} = \underset{\theta}{\operatorname{argmax}} 
			\E \left[ p_{\mathbf{x},\mathbf{y}}(\mathbf{x},y; \theta) \mid \mathbf{y} = y; \, \hat{\theta}^{(n)} 
					\right] ,
\end{equation}
where $\hat{\theta}^{(n)}$ is the guess of the true parameter set on the $n$th iteration. 
It is common to consider the log-likelihood to change the products in the joint likelihood to summations;
the maximizing parameters are the same for either optimization.
The below derivations reference proofs in~\cite[Chap. 13]{Bishop2006} and~\cite{RosenbaumZ:06}.
 
\subsection{Simplification of log-likelihood}

For a collaborative Kalman filter, the log-likelihood simplifies to 
\begin{equation}
	\log L = \log p(x,y; \theta) = \underbrace{\sum_{i=1}^N \log p(x_{i,0}) }_{L_1}
					+ \underbrace{\sum_{i=1}^N \sum_{t=1}^T \log p(x_{i,t} | x_{i,t-1})}_{L_2}
					+ \underbrace{\sum_{i=1}^N \sum_{t=1}^T \log p(y_{i,t} | x_{i,t})}_{L_3} \label{eq:loglikeelements}, 
\end{equation}
with 
\begin{align}
	p(x_{i,0}) 						&\sim \mathcal{N}(x_{i,0}; \, \mu_i, \Sigma_i) ,  \notag \\
	p(x_{i,t}|x_{i,t-1}) 	&\sim \mathcal{N}(x_{i,t}; \, A_{i,t} \, x_{i,t-1}, Q_i) , \notag \\
	p(y_{i,t}|x_{i,t}) 		&\sim \mathcal{N}(y_{i,t}; \, H_{i,t} \, x_{i,t}, R_i) .
\end{align}

Using Fact~\ref{fact:gaussTr}, the first term $L_1$ becomes
\begin{equation}
	\E[L_1] = -\frac{N K}{2} \log(2\pi) - \frac{1}{2} \sum_{i=1}^N \log |\Sigma_i| 
							- \frac{1}{2} \sum_{i=1}^N \tr\left(\Sigma_i^{-1} 		
										\E[(\mathbf{x}_{i,0}-\mu_i)(\mathbf{x}_{i,0}-\mu_i)^T]\right) .
\end{equation}
We then use the identity $\mathbf{x}_{i,0}-\mu_i = (\mathbf{x}_{i,0}- \mathbf{\hat{x}}_{i,0|T}) + (\mathbf{\hat{x}}_{i,0|T} - \mu_i)$ and note that estimation error and innovation of a Kalman filter are uncorrelated to rewrite the expectation of $L_1$ to be
\begin{equation}
	\E[L_1] = c_1 - \frac{1}{2} \sum_{i=1}^N \log |\Sigma_i| 
							- \frac{1}{2} \sum_{i=1}^N \tr\left(\Sigma_i^{-1}
									\left( P_{i,0|T} + (\mathbf{\hat{x}}_{i,0|T} - \mu_i)(\mathbf{\hat{x}}_{i,0|T} - \mu_i)^T 
										\right)\right) .
\end{equation}

We repeat a similar procedure for $L_2$, this time using the identity
\begin{align}
	\mathbf{x}_{i,t} - A_{i,t} \, \mathbf{x}_{i,t-1} 
			&= \mathbf{x}_{i,t} - A_{i,t} \, \mathbf{x}_{i,t-1} 
						+ \mathbf{\hat{x}}_{i,t|T} - \mathbf{\hat{x}}_{i,t|T} 
						+ A_{i,t} \, \mathbf{\hat{x}}_{i,t-1|T} - A_{i,t} \, \mathbf{\hat{x}}_{i,t-1|T} \notag \\
			&= \left(\mathbf{\hat{x}}_{i,t|T} - A_{i,t} \mathbf{\hat{x}}_{i,t-1|T} \right)
						- \left(\mathbf{\hat{x}}_{i,t|T} - \mathbf{x}_{i,t} \right)
						+ A_{i,t} \left( \mathbf{\hat{x}}_{i,t-1|T} - \mathbf{x}_{i,t-1} \right) .
\end{align}
We then rewrite the expectation of $L_2$ as
\begin{align}
	\E[L_2] &= -\frac{N T K}{2} \log(2\pi) - \frac{T}{2} \sum_{i=1}^n \log |Q_i| 
							- \frac{1}{2} \sum_{i=1}^N \sum_{t=1}^T \tr\left(Q_i^{-1}
								\E[(\mathbf{x}_{i,t} - A_{i,t})(\mathbf{x}_{i,t} - A_{i,t})^T]\right) \notag \\
					&= c_2 - \sum_{i=1}^n \frac{T}{2} \log |Q_i|  
								- \frac{1}{2} \sum_{i=1}^N \sum_{t=1}^T \tr \left( Q_i^{-1} \E \left[
							\left\{ \left(\mathbf{\hat{x}}_{i,t|T} - A_{i,t} \mathbf{\hat{x}}_{i,t-1|T} \right)
						- \left(\mathbf{\hat{x}}_{i,t|T} - \mathbf{x}_{i,t} \right)
						+ A_{i,t} \left( \mathbf{\hat{x}}_{i,t-1|T} - \mathbf{x}_{i,t-1} \right) \right\} \right. \right. \notag \\
					&\hspace{.5in} \left. \left. \times \left\{ 
						\left(\mathbf{\hat{x}}_{i,t|T} - A_{i,t} \mathbf{\hat{x}}_{i,t-1|T} \right)
						- \left(\mathbf{\hat{x}}_{i,t|T} - \mathbf{x}_{i,t} \right)
						+ A_{i,t} \left( \mathbf{\hat{x}}_{i,t-1|T} - \mathbf{x}_{i,t-1} \right) \right\}^T \right] \right) \label{eq:l2longform}.
\end{align}
Expanding everything and again noting that the Kalman estimation error and innovation are uncorrelated, 
\eqref{eq:l2longform} simplifies to
\begin{align}
	\E[L_2] &= c_2 - \frac{T}{2} \sum_{i=1}^n \log |Q_i|
			- \frac{1}{2} \sum_{i=1}^N \sum_{t=1}^T \tr \left( Q_i^{-1} \left\{ 
			\left( P_{i,t|T} + \mathbf{\hat{x}}_{i,t|T} \, \mathbf{\hat{x}}_{i,t|T}^T \right) 
			- 2 \left( P_{i,t,t-1|T} + \mathbf{\hat{x}}_{i,t|T} \, \mathbf{\hat{x}}_{i,t-1|T}^T \right) A_{i,t}^T
			\right. \right. \notag \\
			&\hspace{3in} \left.\left.
			+ A_{i,t} \left( P_{i,t-1|T} + \mathbf{\hat{x}}_{i,t-1|T} \, \mathbf{\hat{x}}_{i,t-1|T}^T \right) 
					A_{i,t}^T \right\} \right).
\end{align}

A similar derivation is employed for $L_3$ utilizing
\begin{equation}
	y_{i,t} - H_{i,t} \, \mathbf{x}_{i,t} = (y_{i,t} - H_{i,t} \, \mathbf{\hat{x}}_{i,t|T}) 
								+ H_{i,t} (\mathbf{\hat{x}}_{i,t|T} - \mathbf{x}_{i,t}) .
\end{equation}
Some care is needed in writing out $R_i$ in $L_3$ since $y_{i,t}$ can be of different lengths depending on the observation tensor $O$ and hence only a subset of the noise covariance matrix is required at each time step. To circumvent this issue, we define a fill function that expands the observation vector back to $\mathbb{R}^M$
and a diagonal binary matrix $J_{i,t} \in R^{M \times M}$ with ones in the diagonal positions where 
ratings are observed for user $i$ at time $t$.

Currently, the formulation is extremely general and parameters may change with users and in time.  We can maximize with respect to the log-likelihood but the resulting estimation would be poor and does not exploit the possible similarities between a population of users. To fully realize the benefits of CKF, we make simplifying assumptions that $\mu_i = 0$, $\Sigma_i = \Sigma$, $A_i = A$, $Q_i = Q$, and $R_i = \sigma_R^2 I_M$. 
We now move summations into the trace operator and the log-likelihood simplifies to
\begin{align}
	\E[L_1] &= c_1 - \frac{N}{2} \log |\Sigma| 
							- \frac{1}{2} \tr \left(\Sigma^{-1} \Gamma_1 \right) , \label{eq:l1_final} \\
	\E[L_2] &= c_2 - \frac{N T}{2} \log |Q| - \frac{1}{2} \tr ( Q^{-1} \Gamma_2) , \label{eq:l2_final} \\
	\E[L_3] &= c_3 - \frac{|\mathcal{O}|}{2} \log \sigma_R^2 - \frac{1}{2 \sigma_R^2} \tr (\Gamma_3) , \label{eq:l3_final}
\end{align}
where
\begin{align}
	\Gamma_1 &= \sum_{i=1}^N  \left( P_{i,0|T} + \mathbf{\hat{x}}_{i,0|T} \, \mathbf{\hat{x}}_{i,0|T}^T \right) , \\
	\Gamma_2 &= \sum_{i=1}^N \sum_{t=1}^T \left\{ 
			\left( P_{i,t|T} + \mathbf{\hat{x}}_{i,t|T} \, \mathbf{\hat{x}}_{i,t|T}^T \right)
			- 2 \left( P_{i,t,t-1|T} + \mathbf{\hat{x}}_{i,t|T} \, \mathbf{\hat{x}}_{i,t-1|T}^T \right) A_t^T
			+ A_t \left( P_{i,t-1|T} + \mathbf{\hat{x}}_{i,t-1|T} \, \mathbf{\hat{x}}_{i,t-1|T}^T \right) A_t^T \right\} , \\
	\Gamma_3 &= \sum_{i=1}^N \sum_{t=1}^T \left\{ 
			\fillop(y_{i,t}) \, \fillop(y_{i,t})^T 
			- 2 \fillop(y_{i,t}) \, \mathbf{\hat{x}}_{i,t|T}^T \, V^T
			+ \left(J_{i,t} V \right) \left( P_{i,t|T} + \mathbf{\hat{x}}_{i,t|T} \, \mathbf{\hat{x}}_{i,t|T}^T \right) 
						\left(J_{i,t} V\right)^T \right\} .
\end{align}

\subsection{Determining $\widehat{\Sigma}$, $\widehat{Q}$ and $\widehat{R}$}

To maximize with respect to $\Sigma$, we can differentiate~\eqref{eq:l1_final}, set to zero, and solve. 
Using Facts~\ref{fact:difflogtrace} and~\ref{fact:difftraceinv}, 
\begin{equation}
	\frac{\partial \E[L_1]}{\partial \Sigma} = - \frac{N}{2} \left( \Sigma^{-1} \right)^T + \frac{1}{2} \left( \Sigma^{-1} \Gamma_1 \Sigma^{-1} \right)^T ,
\end{equation}
and solving gives
\begin{equation}
 	\widehat{\Sigma} = \frac{1}{N}\Gamma_1 . 
\end{equation}

If we had further assumed that $\Sigma = \sigma^2_U I_K$, 
then~\eqref{eq:l1_final} would simplify to
\[ \E[L_1] = c_1 - \frac{N K }{2} \log \sigma^2_U - \frac{1}{2 \sigma^2_U} \tr \left( \Gamma_1 \right) , \]
and maximization yields~\eqref{eq:varU}.

The derivations for $\widehat{Q}$ and $\widehat{R}$ follow similarly and lead to
\eqref{eq:varQ} and \eqref{eq:varR} respectively.

\subsection{Determining $\widehat{A}$ and $\widehat{V}$}

Rewriting~\eqref{eq:l2_final} as
\begin{equation}
	\E[L_2] = c_A + \tr ( Q^{-1} A_2 A^T) - \frac{1}{2} \tr ( Q^{-1} A A_1 A^T) , 
\end{equation}
where $c_A$ is the collection of terms that do not depend on $A$, 
we maximize using the same procedure as for $\widehat{\Sigma}$. 
We utilize Facts~\ref{fact:difftrace1} and~\ref{fact:difftrace2} while noting that $A_1$, $A_2$ and $\widehat{Q}$ are symmetric and invertible, and the maximization yields \eqref{eq:A}.

Following a similar procedure for optimization of $\widehat{V}$, we express \eqref{eq:l3_final} as
\begin{equation}
	\E[L_3] = c_V + \frac{1}{\sigma^2_R} \tr \left( V_2 V^T \right) 
				- \frac{1}{2 \sigma^2_R} \tr \left( \sum_{i=1}^N \sum_{t=1}^T J_{i,t} V V_{i,t} V^T J_{i,t}^T \right). 
\end{equation}
In this case, $V$ cannot be expressed as a matrix product, but each row can. 
Noting $J_{i,t} = J^T_{i,t}$ and $J_{i,t} J_{i,t} = J_{i,t}$, the maximization over each row yields \eqref{eq:V}.

\bibliographystyle{IEEEtran}
\bibliography{IEEEabrv,matrixfact}

\end{document}